\documentclass{article}

\usepackage[final]{corl_2020} 

\usepackage{wrapfig}
\usepackage{subcaption}
\usepackage{graphicx}
\usepackage{amsmath}
\usepackage{amsfonts}
\usepackage{mathtools}
\usepackage{array}
\usepackage{booktabs}
\usepackage{makecell}
\usepackage{multirow}
\usepackage{caption}
\usepackage{gensymb}
\usepackage{amssymb}
\usepackage{cleveref}

\crefformat{footnote}{#2\footnotemark[#1]#3}

\title{Transformers for One-Shot Visual Imitation}

\author{
  Sudeep Dasari\\
  Robotics Institute \\
  Carnegie Mellon University, USA \\
  \texttt{sdasari@andrew.cmu.edu} \\
   \And
   Abhinav Gupta \\
   Robotics Institute \\
   Carnegie Mellon University, USA \\
   \texttt{gabhinav@andrew.cmu.edu} \\
}

\begin{document}
\maketitle


\begin{abstract}
    Humans are able to seamlessly visually imitate others, by inferring their intentions and using past experience to achieve the same end goal. In other words, we can parse complex semantic knowledge from raw video and efficiently translate that into concrete motor control. Is it possible to give a robot this same capability? Prior research in robot imitation learning has created agents which can acquire diverse skills from expert human operators. However, expanding these techniques to work with a single positive example during test time is still an open challenge. Apart from control, the difficulty stems from mismatches between the demonstrator and robot domains. For example, objects may be placed in different locations (e.g. kitchen layouts are different in every house). Additionally, the demonstration may come from an agent with different morphology and physical appearance (e.g. human), so one-to-one action correspondences are not available. This paper investigates techniques which allow robots to partially bridge these domain gaps, using their past experience. A neural network is trained to mimic ground truth robot actions given context video from another agent, and must generalize to unseen task instances when prompted with new videos during test time. We hypothesize that our policy representations must be both context driven and dynamics aware in order to perform these tasks. These assumptions are baked into the neural network using the Transformers attention mechanism and a self-supervised inverse dynamics loss. Finally, we experimentally determine that our method accomplishes a $\sim 2$x improvement in terms of task success rate over prior baselines in a suite of one-shot manipulation tasks.\footnote{For code and project video please check our website:  \url{https://oneshotfeatures.github.io/}}
\end{abstract}


\section{Introduction}
\label{sec:intro}

Imitation is one of the most important cornerstones of intelligence. Watching other humans act, inferring their intentions, and attempting the same actions in our own home environments allows us to expand our skill set and enhance our representations of the world~\citep{vogt2007visuo}. On the other hand, robots - while capable of imitating skills like table tennis~\citep{mulling2013learning} and driving~\citep{pomerleau1989alvinn} -- are much less flexible when it comes to visual imitation. Most prior work in robotic imitation assumes that the agent is trying to acquire a single skill from demonstration(s) collected kinesthetically~\citep{pastor2011online} (i.e. a human manually guides a robot) or via tele-operation~\citep{zhang2018teleop}. These approaches can work so long as the target test-time task and environment are do not significantly differ from those seen during training. Is it possible to develop a robotic agent which can learn to imitate without these restrictions?

\begin{figure}[t]
    \centering
    \includegraphics[width=0.9\textwidth]{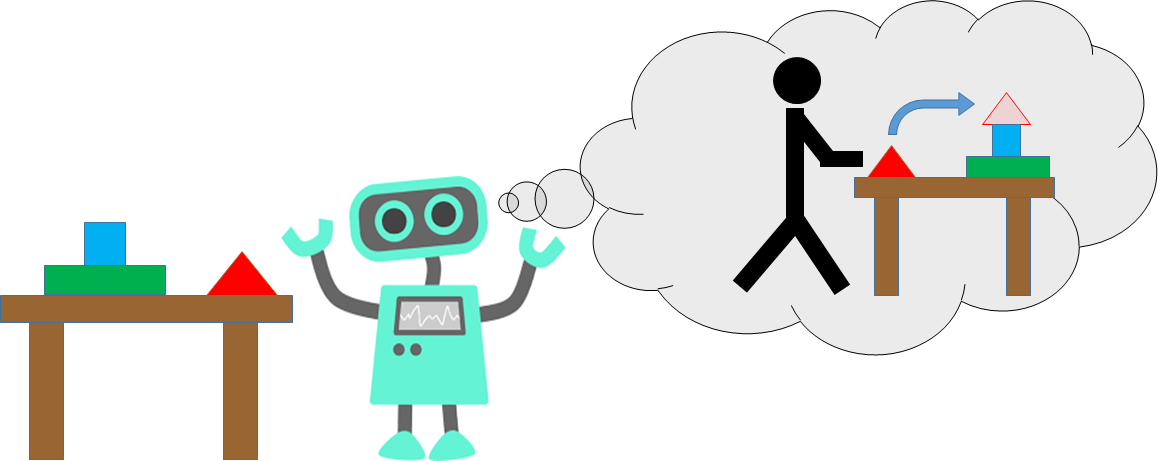}
    \caption{What should the robot do given video from another demonstration agent? A human would immediately know to place the red triangle on the blue square, and can use their past experience to execute the task. Is it possible to teach a robot to do the same?
    }
    \label{fig:oneshot_description}
    \vspace{-0.25in}
\end{figure}

Visual imitation requires extracting a higher level goal from the visual demonstration and using the inferred goal to predict actions from pixels. But how does one represent goal/intention and how can this contextual information be incorporated into the policy function itself? There are three primary approaches in prior work: the first approach is to represent goals/intentions as pixels by generating goal images, and then inferring actions given current observations and inferred goals~\citep{sharma2019third, smith2019avid}. While this approach is intuitive and interpretable, it is difficult to generate pixels, in a way that respects structural differences in the image. Figure~\ref{fig:oneshot_description} shows an example with well defined task semantics, but where a change in object positions makes it difficult to visually map the human state to the robot environment. The second approach has been to model visual imitation as a one-shot learning problem~\citep{duan2017one}, which can be solved with meta-learning algorithms~\citep{finn2017one}. Here, a robot is given a single example, in the form of a video or demonstration (e.g. video + control telemetry), and must use that information to perform new instances of the same task. The demonstration is used to update the parameters of a policy function and the updated policy is executed on the robot. Domain gaps can be addressed with a learned adaptive los function~\citep{DAML}. While the one-shot formalism is very useful, estimating policy parameters from a single example can be an extremely difficult problem and prone to over-fitting. 

In this paper, we explore a third alternative: task-driven features for one-shot learning. We process both observations from the target agent and demonstrations frames from a "teacher" agent in order to extract context-conditioned state representations.
What neural network architectures can create task-driven features? While in the past, approaches such as LSTMs have been used, in this work, we focus on self-attention architectures. In particular, the Transformers architecture - while simple - has seen broad success in  NLP~\cite{vaswani2017attention} and Vision~\cite{wang2018non} tasks. Furthermore, using attention for control tasks has has basis in biology and psychology. Indeed, humans use attention mechanisms to create context driven representations~\citep{rothkopf2007task}, and directly supervising policies with human attention can dramatically increase task performance~\citep{zhang2018agil}. 

In this paper, we propose using transformers~\cite{vaswani2017attention} (or non-local self-attention modules~\cite{wang2018non}) to extract relational features which act as input state vectors for the policy function. Our transformers take as input both the spatial ResNet Features from teacher demonstration and the target agent. This allows the policy to automatically adapt its features to the task at hand, by using context frames to focus only on important task-specific details. For example, in Figure~\ref{fig:oneshot_description} the robot could use human context frames to focus only on relevant details like the red block's location, and entirely ignore distracting elements like the table's leg. However, transformer features could easily end up improperly weighting important details during test time. We propose to solve this issue by further supervising the state representation learning with an unsupervised inverse dynamics loss. This loss constrains the learning problem and ensures the final representations can model the underlying dynamics, as well as task specific details. Ultimately, our method achieves significant improvements over one-shot imitation learning baselines on a suite of pick and place tasks: our final policies demonstrate a $2$x performance gain and can match baseline performance with $3$x fewer data-points.

\section{Related Work}
\label{sec:relatedwork}
Learning from Demonstration (LfD) is a rich and diverse field of study which focuses on enabling robots to learn skills from human or other expert demonstrations. A thorough review is out of scope for this paper, so we gladly refer the reader to survey articles~\citep{argall2009survey, billard2008survey, schaal1999imitation}. Of prior work, Behavior Cloning (BC)~\citep{ross2011reduction, bain1995framework}, a common formulation of LfD, is most related to our project. BC involves imitating an expert agent given a set of trajectories (a.k.a time series of observations and actions), by fitting a function which approximates the expert's action in a given state. This simple formulae has proven successful in imitating a wide range of behaviors from visual inputs, including robotic manipulation tasks~\citep{rahmatizadeh2018vision} and driving~\cite{bojarski2016end}. These methods have been extended to situations where expert observations are present without action labels~\cite{torabi2018behavioral}, including prior work which linked this problem to inverse dynamics minimization~\cite{yang2019imitation}. However, both of these approaches require the demonstration agent match the imitator.

BC algorithms often assume that they are approximating a single state conditioned policy. In an environment with multiple tasks or multiple variations of the same task, this constraint can be limiting. Work on goal conditioned imitation learning seeks to relax these assumptions by allowing for policies which condition on a goal variable alongside the current state, and adjust their behavior accordingly. There are myriad ways to introduce goal conditioning, including with the robot's state~\citep{ding2019goal}, "goal" images of the final state~\citep{mandlekar2019iris, lynch2020learning, fu2018variational}, natural language~\citep{lynch2020grounding}, and video or images of humans~\citep{lin2020concept, xie2019improvisation}. In our project, we assume the robot has a single video of another agent (be it another robot or a human) doing a task, and must complete that same task itself using past experience. This is a specific instance of the one-shot learning problem~\citep{duan2017one}, and has been investigated before previously using meta-learning with an adaptive loss~\citep{DAML}. Instead of using meta-learning, we propose to attack this problem with an attention mechanism over image frames.

A challenge in this line of work is learning visual representations which can enable the robot to deduce the task from video of another agent \emph{and} perform the task itself. Work in computer vision demonstrated that deep neural networks are capable of learning such flexible representations for action recognition~\citep{simonyan2014two} and state estimation~\citep{lee2019camera}, but often require large image datasets to fully train. Unfortunately, collecting ImageNet~\citep{deng2009imagenet} scale datasets on robotics platforms is prohibitively expensive, due to the cost of continuous robot operation and hardware fragility. Work in self-supervised learning ~\citep{he2020momentum, chen2020simple, grill2020bootstrap} offers a glimmer of hope, by showing how large and (relatively) cheap sets of unlabelled images can be used to learn expressive and useful representations for other downstream tasks. These representations could be used directly as reward functions~\citep{sermanet2016unsupervised,sermanet2018time}, but it can be very difficult to define rewards for a suite of tasks. Instead, unsupervised learning techniques alongside simple data augmentation can be used to increase data efficiency when directly acquiring policies with reinforcement learning~\citep{srinivas2020curl, laskin2020reinforcement, kostrikov2020image}. Even simpler self-supervised losses - like inverse modelling (i.e. predicting action between two sequential states) - can be used to learn robust policies which adapt to new environments~\citep{hansen2020self}. Our goal in this project is to apply these insights in representation learning to the one-shot imitation learning problem.

\section{Our Method}
\subsection{Problem Definition}
\label{sec:method}

Our method follows prior work~\citep{finn2017one,DAML}, and formalizes the one-shot imitation learning problem as supervised behavior cloning on a data-set of tasks. For each task $\mathcal{T}$ (e.g. place blue bottle in bin), we have several demonstration videos and target trajectories. Note that the demonstration videos and target trajectories are semantically similar tasks but could have different starting/end states. We represent each demonstration video as $v_i$ and each target trajectory, $t_i$, as a temporal sequence of observations ($o$) and actions ($a$). Hence, $t_i = \{ (o_i^{(1)}, a_i^{(1)}), \dots, (o_i^{(k)}, a_i^{(k)} )\}$. 

Models are trained on a dataset of tasks $\mathcal{D} = \{ \mathcal{T}_1, \dots, \mathcal{T}_n \}$. During test time, new test tasks - $\mathcal{T}_{test}$ - are sampled which the model must successfully control the imitator agent to perform. Thus, all methods are evaluated on task success rates in held out environments. Our setup is challenging because: (a) morphological differences between demonstration and target agent (e.g. one is human and other is robot arm); (b) missing correspondence between demonstration videos and target trajectories.

\begin{figure}[t]
    \centering
    \includegraphics[width=1\textwidth]{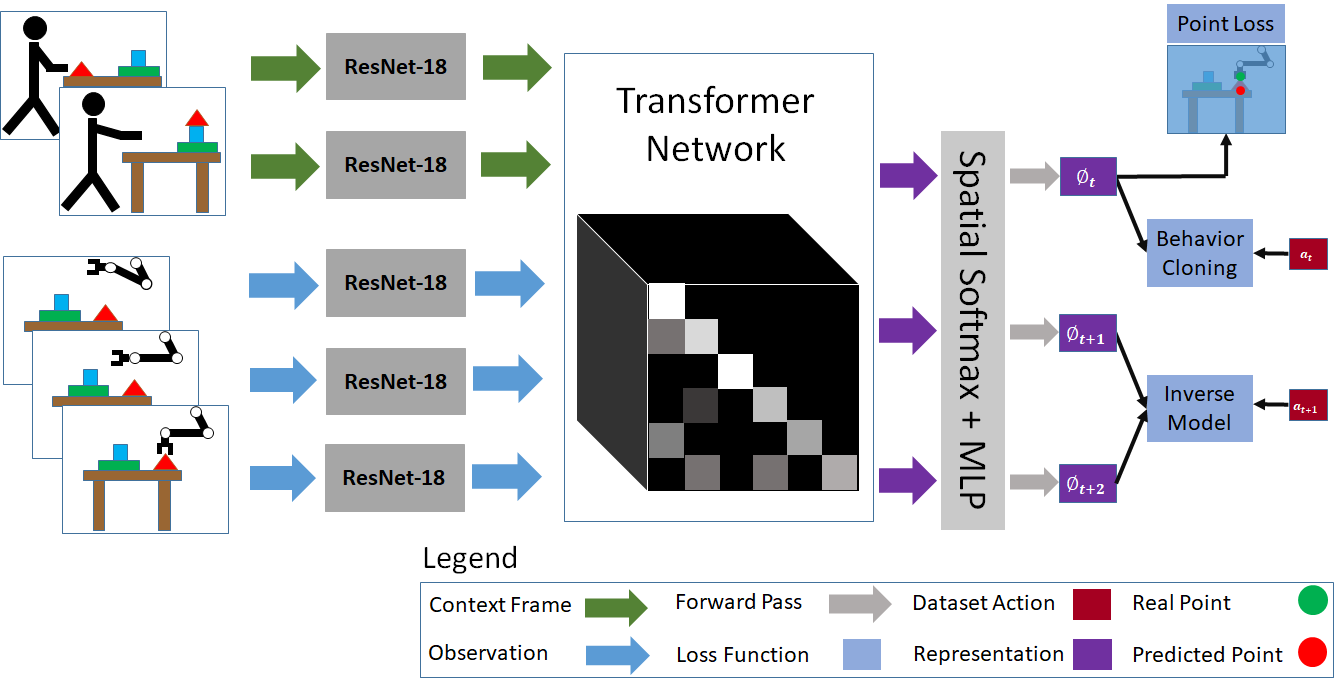}
    \caption{Our method uses a Transformer neural network to create task-specific representations, given context and observation features computed with ResNet-18 (w/ added positional encoding). The attention network is trained end-to-end with a behavior cloning loss, an inverse modelling loss, and an optional point loss supervising the robot's future pixel location in the image. 
    }
    \label{fig:model}
    \vspace{-0.1in}
\end{figure}

\subsection{Feature Learning with Transformers}
\label{sec:methodattn}
Given video context from a demonstrator agent and image frames from the test environment, our representation module must deduce relevant features and efficiently pass them on to later stages of the pipeline for action prediction. For example, when given a video of a green bottle being dropped in a bin, the vision module should detect and represent the green bottle in its own environment while ignoring other distracting objects. We propose to learn this mechanism end-to-end using self-attention Transformer modules~\citep{vaswani2017attention}, in the hope that this powerful inductive bias helps the policy perform tasks successfully. 

Before the attention module, individual images from both the context video and current state are passed through a ResNet-18 architecture~\citep{he2016deep}, and spatial features (size $[512, T, H, W]$) are collected before the average pooling step. At this stage, the features are flattened (size $[512, T*H*W]$) and sinusoidal positional encodings~\citep{vaswani2017attention} are added to the tensor (i.e. time and space treated as single dimension). These embeddings can allow neural networks to represent higher frequency functions~\citep{tancik2020fourier}, and we empirically found that they were crucial to preserving spatial and temporal information in the attention module. After adding positional encodings, the features are reshaped to their original size.

Next, the non-local multi-head attention operator is applied to the input tensor. We adopt a specific implementation of the Transformers self-attention module presented in~\citet{wang2018non}, which we augment with multi-headed self-attention~\citep{vaswani2017attention}. First, the module generates \textbf{K}ey, \textbf{Q}uery, and \textbf{V}alue tensors by applying three separate 3D spatio-temporal convolutions (we use kernel size $k=1$) with ReLU activation to the input tensor. To be clear, each convolution layer's input and output are $[d, T, H, W]$ tensors, where $d$ is the Transformer's embedding size. These generated key, query, and value tensors are then flattened and projected down $n$ separate times - once for each attention ``head" - before attention is applied (final shape per head $[d, T*H*W]$). The self-attention operator is applied to each head individually. Considering attention head $j$, temperature parameter $\tau$, and projected tensors $K_j, Q_j, V_j$, this amounts to:
$$ A_j = \textbf{softmax}(K_j^T Q_j / \tau) \hspace{8mm} V_j^{(out)} = V_j A_j$$
The individual attention heads are then concatenated together channel-wise, and then projected back to the original $512$ dimension size with another 3D convolution ($O = \textbf{Conv3D}(\textbf{concat}[V_1^{(out)}, \dots, V_n^{(out)}])$). Note that this multi-head attention operator can be implemented with little overhead using batched matrix multiplication. Dropout~\citep{srivastava2014dropout}, then a residual connection, and finally batch normalization~\citep{ioffe2015batch} are applied to get the final output $f(x) = \textbf{batchnorm}(x + \textbf{dropout}(O))$, with final size $[512, T, H, W]$. In order to appropriately apply this to behavior cloning (where $o_{t+1}$ is not known during test time), we make this operation causal by appropriately padding the 3D convolution operators and masking the attention. 

\subsection{Goal Conditioned Behavior Cloning}
As discussed previously, our objective is to learn a policy $\pi(a_t | o_{1:t}, v)$ which ingests the current (or optionally all previous) state observations alongside a context video, and predicts a distribution over possible actions the expert policy would select. We process the input video stream with stacked attention modules to yield fixed size spatial features, with one feature map per time-step. The features are projected down to a fixed size representation vector using a spatial softmax operator~\citep{levine2016end}, followed by a multi-layer perceptron with ReLU activations, and finally L2 normalization to unit length. This representation $\phi_t = F(o_{1:T}, v)$ is used for action prediction.

\textbf{Multi-Modal Action Prediction:} One of the most naive ways to predict $\pi(a_t | o_{1:t}, v)$ from $\phi_t$ is to simply parameterize the policy as a normal distribution $\pi(a_t | o_{1:t}, v) = \mathcal{N}(\mu(\phi_t), \sigma(\phi_t))$, and to sample actions from that. However, this approach can run into severe limitations when the real expert distribution is multi-modal. Consider a robot attempting to top-down lift a cup by its handle. Rotating the gripper by 90\degree or -90\degree, but not rotating at all (i.e. the mean action) would result in task failure since the gripper would close on top of the handle. Prior work~\citep{rahmatizadeh2018vision, rahmatizadeh2018virtual, lynch2020learning} showed this limitation matters in practice, and rectifies the situation by predicting a mixture of uni-modal distributions. We adopt the same solution used by \citet{lynch2020learning}. First, we discretize the action space (discussed in detail in Section~\ref{sec:sim_env}) and then parameterize the policy as a discretized logistic mixture distribution~\citep{salimans2017pixelcnn++}. For each timestep, we predict $k$ logistic distributions with separate mean and scale, and form a mixture by convexly weighting them with vector $\alpha$. The behavior cloning training loss is simply negative log-likelihood for this distribution:
$$ \mathcal{L}_{BC}(\mathcal{D}, \theta) = - \ln(\Sigma_{i=0}^k \hspace{1mm} \alpha_k(\phi_t) \hspace{1mm} P(a_t, \mu_i(\phi_t), \sigma_i(\phi_t))$$

Where, $P(a_t, \mu_i(\phi_t), \sigma_i(\phi_t)) = F(\frac{a_t + 0.5 - \mu_i(\phi_t)}{\sigma_i(\phi_t)}) - F(\frac{a_t - 0.5 - \mu_i(\phi_t)}{\sigma_i(\phi_t)})$ and $F(\cdot)$ is the logistic CDF. During test time, actions are simply sampled from the distribution and executed on the robot without rounding. For most of our experiments, the model performed best when using two mixture components and learned constant variance parameters per action dimension.

\subsection{Inverse Model Regularizer}
Our method also adds a self-supervised inverse modeling objective to act as a regularizer to the behavior cloning loss during training. Context and trajectory snippets are sampled from the dataset, and images in them are randomized with sampled translations, color shifts, and crops. This randomization is applied consistently to frames from the context video, whereas images from the agent's observation stream (a.k.a trajectory images) are randomized \textit{individually}. This randomized image stream is passed through the attention and representation modules to generate $\Tilde{\phi_t}$. The representations $\Tilde{\phi}_t$ and $\Tilde{\phi}_{t+1}$ are used to predict a discretized logistic mixture distribution over intermediate actions. Thus, the inverse loss is:
$$ \mathcal{L}_{INV}(\mathcal{D}, \theta) = - \ln(\Sigma_{i=0}^k \hspace{1mm} \alpha_k(\Tilde{\phi}_{t}, \Tilde{\phi}_{t+1}) \hspace{1mm} \text{logistic}(\mu_i(\Tilde{\phi}_{t}, \Tilde{\phi}_{t+1}), \sigma_i(\Tilde{\phi}_{t}, \Tilde{\phi}_{t+1})))$$
We share parameters between the behavior cloning and inverse modeling objectives for the attention module, representation module, and distribution prediction heads (i.e. after first layer). In practice, we use the randomized image stream for both tasks as well, in order to minimize memory consumption.

\subsection{Point Prediction Auxiliary Loss}
Finally, our model uses $\phi_t$ to predict a 2D keypoint location corresponding to the location of the gripper in the image $H$ timesteps in the future. Ground truth for this auxiliary loss is easy to acquire given either a calibrated camera matrix or object detector trained on the robot gripper. One could instead predict the 3D gripper position in world coordinates if neither is available. While not strictly needed for control, this loss is very valuable during debugging, since it lets us visually check during training if the model understand where the robot ought to be $H$ timesteps in the future. The point prediction is parameterized with a simple multi-variate 2D normal distribution $\hat{p}_{t+H} \sim \mathcal{N}(\mu(\phi_t), \Sigma(\phi_t))$ with loss $\mathcal{L}_{pnt}(\mathcal{D}, \theta) = - \ln (\text{likelihood}(p_{t+H}, \hat{p}_{t+H}))$. Thus, the overall loss for our method is:
$$ \mathcal{L}(\mathcal{D}, \theta) = \lambda_{BC} \hspace{1mm} \mathcal{L}_{BC}(\mathcal{D}, \theta) + \lambda_{INV} \hspace{1mm} \mathcal{L}_{INV}(\mathcal{D}, \theta) + \lambda_{pnt} \hspace{1mm} \mathcal{L}_{pnt}(\mathcal{D}, \theta) $$

\section{Experimental Results}
\label{sec:result}
Our model is evaluated on robotic manipulation tasks - namely pick and place tasks - in simulation using multi-agent MuJoCo~\citep{todorov2012mujoco} environments. Our evaluations investigate the following questions: (1) can our model perform new task instances (defined in~\ref{sec:sim_env}) previously unseen during training? And (2) what components (e.g. inverse loss, etc.) are most crucial for successful control?

\begin{wrapfigure}{l}{.35\columnwidth}
    
    \centering
    \includegraphics[width=0.35\columnwidth]{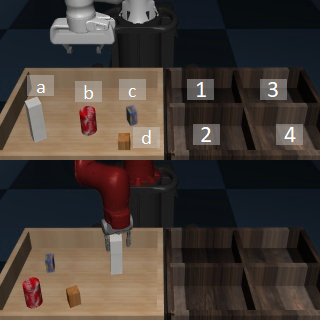}
    \caption{Our base environment is adopted from RoboTurk~\citep{mandlekar2018roboturk}. The $16$ tasks consist of taking an object (a-b) to a bin (1-4). Top robot is agent and bottom is demonstrator.
    }
    \label{fig:base_task}
    \vspace{-0.1in}
\end{wrapfigure}

\subsection{Simulation Environment and Tasks}
\label{sec:sim_env}
\textbf{Environment Description:} The environments we use are modified variants of those originally presented in RoboTurk~\citep{mandlekar2018roboturk}. Visually, the \textit{base environment} - shown in Figure~\ref{fig:base_task} - is the exact same as the original from RoboTurk, except the object meshes are replaced with primitive geometric types (e.g. boxes and cylinders) in order to improve simulation contact stability and run-time. This modification results in only minor visual differences. In order to investigate visual imitation across agent morphology, we use duplicate versions of the environment with two visually distinct robots. The Sawyer robot (red robot in Figure~\ref{fig:base_task}) provides demonstration videos and the Panda robot (white robot in Figure~\ref{fig:base_task}) acts as the agent which our model must control. Both environment's action spaces are modified to support end-effector control. Given a target $x,y,z$ position, rotation in axis-angle form, and gripper joint angle the environment solves for desired robot joint angles with inverse kinematics and sends joint velocities to the robot using a simple PD controller. Thus, the final action space consists of a target pose discretized into 256 independent bins per dimension in order to support our behavior cloning loss. It's important to note that the demonstrations we train on do not cover the whole state space, so the robot is mostly constrained to 3-DOF movement. 

\textbf{Task Definition:} A ``task instance" consists of picking an object from a specific start location - uniformly distributed on the table in Fig.~\ref{fig:base_task} - and placing the object in one of the four bins on the right. Task instances are grouped into ``tasks" based on shared properties. For example, picking a milk carton (from Fig.~\ref{fig:base_task}) and placing it into bin 1 is a task, and different task instances are constructed by changing the carton's start position. This precise definition allows us to collect a suite of train task instances, train models on that data, and test generalization to new task instances.

\textbf{Data Collection Methodology:} Training data is collected using an expert pick-place policy (built using privileged information from the simulator) in the target environment(s). For each task ($\mathcal{T}$) we repeatedly, sample a demonstration video ($v_i$) by executing the expert policy on the Sawyer robot, then shuffle the objects, and sample an expert trajectory ($t_i$) by executing the expert policy on the Panda robot. This way a dataset of tasks is formed from individual task instances.

\subsection{Baseline Comparisons}
\label{sec:baseline}
\begin{figure}

    \centering
    \footnotesize
\begin{tabular}{ |l|c|c|c| } 
 \toprule \thead[l]{\textbf{Model}}
   & \thead{Reaching Success\!\!} & \thead{Picking Success\!\!} & \thead{Placing/Overall Success\!\!} \\
 \hline
 Our Method  &  $\textbf{99.4\%} \pm 1.2\%$ &  $\textbf{92.5\%} \pm 4.1\%$ &  $\textbf{88.8\%} \pm 5.0\%$\\
 \hline
 Contextual-LSTM  &  $38.8\% \pm 7.6\%$ &  $26.3\% \pm 6.9\%$ &  $23.8\% \pm 6.7\%$ \\
  \hline
 DAML~\citep{DAML}  &  $36.9\% \pm 7.6\%$ &  $10.6\% \pm 4.8\%$ &  $6.9\% \pm 4.0\%$ \\
  \hline
 DAML Auxiliary  &  $47.8\% \pm 7.4\%$ &  $17.8\% \pm 5.6\%$ &  $13.3\% \pm 5.0\%$ \\
 \bottomrule
\end{tabular}
 \captionof{table}{Comparison between our method and baselines in $16$ pick and place tasks. Values indicate success rates and $95\%$ confidence intervals for ``stages''\cref{stagenote} in the overall pick and place task.
 \vspace{-0.4cm}
\label{tbl:baseline_table}}
\end{figure}

Our investigation begins by evaluating our method's performance in $16$ tasks in the base environment (Figure~\ref{fig:base_task}). We seek to determine the robot's physical competency at manipulating all four objects, as well as its ability to deduce which task it should perform from context video. A natural way to quantify this is by breaking down the $16$ pick and place tasks into ``reach," ``pick," and ``place" stages\footnote{\label{stagenote}Reaching is defined as placing the gripper within $< 0.03$ units from the target object, picking requires stably lifting the object $> 0.05$ units off ground, and placing requires putting the object in the correct bin (a.k.a task completion)}, and reporting success rates on each stage individually. Failure modes can be successfully deduced from these rates. For example, since reaching is a physically easy task, if the robot does not reach the object then it is likely unable to deduce the target object from the context frames. Furthermore, if the robot reaches the object but is unable to pick it up, its physical dexterity (or lack thereof) is likely to blame.

We collect $100$ train task instances using the methodology described previously for each of the $16$ tasks. That amounts to $1600$ total demonstration videos alongside $1600$ expert robot trajectories. We train our method on the dataset and compare against the following baselines:

\begin{itemize}
    \item \textbf{Contextual-LSTM:} This baseline utilizes a standard Encoder-Decoder LSTM~\citep{hochreiter1997long, sutskever2014sequence} (augmented with self-attention~\citep{bahdanau2014neural,cheng2016long}), to first consume the context video, and then predict actions from encoded observations. It uses the same mixture distribution our model uses. Before LSTM processing, images frames are embedded using a pre-trained ResNet-18~\citep{he2016deep} neural net combined with spatial-softmax~\citep{levine2016end} and fully-connected layers.  The whole network is trained end-to-end with a behavior cloning loss.
    \item \textbf{Domain Adaptive Meta-Learning:} DAML~\citep{DAML} uses a learned loss function to adapt a neural network's parameters to perform the desired task. We used a wider version of the network used in the original paper, since we found that using deeper models (like ResNet-18) resulted in overfitting on this task. To increase performance, the same discrete logistic action distribution is used. DAML is trained end-to-end with the MAML meta-learning algorithm~\citep{finn2017model} using a behavior cloning loss, along with explicit supervision of the pick and drop locations.
    \item \textbf{DAML-Auxiliary:} This method uses the same meta-learning model described above, except only the predicted pick and place locations are used during test time. Given this prediction, a grasp motion is executed in the environment using a hard coded grasp policy.
\end{itemize}

For each of the $16$ tasks, the models are prompted to perform new task instances (unseen during training) using freshly generated context videos. Success rates for our method and baselines (averaged across tasks) are shown in Table~\ref{tbl:baseline_table}. As you can see, our method is the only one which can reliably perform new task instances. Its overall success rate is double the competing models' reaching success rate, including the DAML-auxiliary model which makes strong task assumptions, and the LSTM model which uses embedding level attention. The LSTM baseline's (which uses standard attention) relative failure supports our hypothesis that the Transformer architecture uniquely enables difficult visual processing. For additional experiments testing generalization to new objects (i.e. new tasks instead of new task instances) refer to Appendix~\ref{app:obj}.


\subsection{Architecture Ablation}
\label{sec:AA}
While the our model clearly outperforms the other baselines, it is unclear if the Transformers architecture or additional losses deserve more credit. To test this thoroughly, the Transformers model is tested against an ablated version of itself without the attention mechanism (i.e. just temporal-spatial convolutions) using the same base environment comparison described before. Furthermore, models are trained with various versions of the baseline neural network architectures, alongside the additional loss terms. Specifically, $4$ baseline architectures are considered: $2$ of them adopt the small convolutional network used in prior work~\citep{DAML,zhou2019watch} either with or without an additional LSTM~\citep{hochreiter1997long} on top, and the other $2$ use ResNet features~\citep{he2016deep} (again with or without LSTM). Note all architectures were tuned to maximize \textit{their own} test performance rather than to match some other metric (e.g. number of parameters), since doing so often led to worse results for the baseline (e.g. larger LSTMs overfit more than Transformers). Results are presented in Figure~\ref{fig:ablate_trans}. The key takeaways are encouraging. First, the Transformers architecture (w/ attention) outperforms a library of other architectures for this task by large margins, even using the same losses. Furthermore, the baselines perform better when trained with the additional losses compared to being trained purely with a behavior cloning loss as done before (contextual-LSTM's success rate improves $20\% \rightarrow 40\%$).

\begin{minipage}[t]{\textwidth}
    \begin{minipage}[b]{0.48\textwidth}
	    \centering
	    
        \includegraphics[width=1.0\columnwidth]{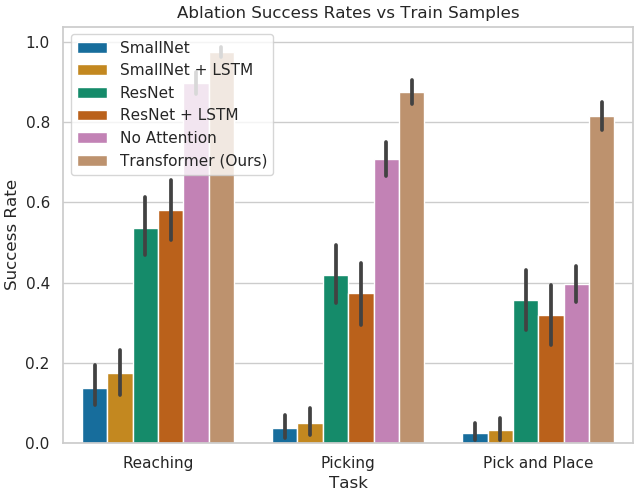}
	    \captionof{figure}{Our Transformer model is compared against other neural networks (all trained w/ our losses and code) to determine how useful the attention mechanism really is. The Transformer architecture outperforms all others, including a version of itself w/out attention.}
	    \label{fig:ablate_trans}
    \end{minipage}
    \hfill
    \begin{minipage}[b]{0.48\textwidth}
        \centering
        \includegraphics[width=1.0\columnwidth]{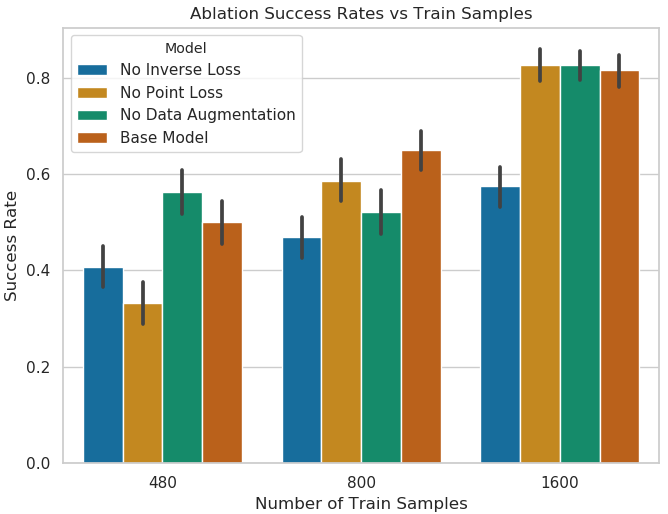}
        
        \captionof{figure}{We compute success rate v.s number of train samples for our method and versions with one loss excluded (all w/ Transformer). Note the model without inverse loss is usually outperformed when compared to its peers trained on the same data.}
        \label{fig:ablate_exp}
    \end{minipage}
    \vspace{0.1cm}
\end{minipage}

\vspace{-0.12in}
\subsection{Loss Function and Method Ablations}
\label{sec:LFA}

        


Given that our training losses/code boosted baseline architecture performance compared to using just behavior cloning, we now seek to test exactly which component was most useful. It's entirely possible that some of the additional parts offer more utility in the "low-data" regime where over-fitting is more likely, and thus are less useful when more data is present. Thus, we collect two more versions of the base environment dataset with fewer samples ($480$ and $800$ samples pairs), and train three ablations - one model without the inverse loss, one without the point loss, and one without data augmentation - alongside our base model on all three datasets (two new sets + original). That results in a total of 12 models, all of which we evaluate in the same manner as before. Overall success rates for all models are in Figure~\ref{fig:ablate_exp}. Note that the model without the inverse loss is outperformed by its counterparts in two out of three datasets, whereas the point loss only makes a significant difference in the smallest dataset. Indeed as the number of datapoints increases, so does the importance of the inverse loss: the model without inverse loss is more than $~25\%$ worse than its counterparts in the $N=1600$ case! While the inverse loss clearly makes a difference, this cannot be observed as ``positive transfer" in the behavior cloning train/test loss (see Appendix~\ref{app:loss}). This suggests inverse loss regularization helps test time performance in ways not captured in the training objective. Finally, conditioning our policy on context video proved to be more effective than just feeding it the last frame, which indicates the demonstration helps our model determine which task to perform compared to using a ``goal image" frame. For more check Appendix~\ref{app:tstep}.

\section{Discussion}
In this project we explore the one-shot visual imitation learning problem. Our experiments highlight two technical contributions - applying the Transformers architecture to one-shot imitation tasks and a self-supervised inverse modelling objective - which both result in large performance gains over baseline one-shot imitation learning approaches. More specifically, our ablations show that our model trained without the self-supervised inverse loss performs significantly worse when compared to other versions with the inverse loss, and all of our Tansformers models (even without inverse loss) outperform a Seq2Seq LSTM trained with traditional ``embedding level'' attention mechanisms by roughly $2$x.

The main takeaway here is that injecting the right biases - both in terms of network design and the loss function - can help policies perform better during test-time. We believe that the Transformer's attention mechanism provides such a bias by allowing for task conditioned representations, whereas the inverse model forces the policy to preserve information which is needed for robust control during test time. We hope that these findings prove useful to others working on one-shot imitation learning and goal conditioned reinforcement learning in general.

\clearpage
\acknowledgments{We'd like to begin by acknowledging the students and collaborators at CMU who gave valuable feedback which made the final paper much stronger. In particular, we'd like to recognize Sam Powers, Devendra Chaplot, Kenneth Marino, Adithya Murali, and Shubham Tulsiani. Finally, this research was funded by ONR MURI, the ONR Young Investigator Award to Abhinav Gupta and the DAPRA MCS award.}


\begin{footnotesize}
\bibliography{references}  

\end{footnotesize}

\newpage
\appendix

\section{Appendix}

\subsection{Baseline Comparisons: Multi-Object Environments}
\label{app:obj}
While the prior experiments showed our model could successfully generalize to new task instances, can it also generalize to new tasks including unseen objects? To answer this question the baseline comparisons (described in Section~\ref{sec:baseline}) are repeated in environments with multiple objects. Importantly, the objects used during test time are unseen during training.

\textbf{Environment Description:} The \textit{multi-object environment} is cloned from the base environment (presented in Section~\ref{sec:sim_env}) and modified to include more objects with different shapes and textures. Note that while object appearance and shape is randomized, dynamical properties - like friction - are kept constant since they cannot be visually judged. The simulator has $30$ unique objects, $26$ of which are seen during training and $4$ are only used during test time.

\textbf{Data Collection Process:}  To collect train tasks, $4$  objects are sampled from the $26$ train objects, which results in an environment with $16$ tasks. For each task, multiple task instances composed of expert demonstration videos ($v_i$) and imitator trajectories ($t_i$) are collected using the same methodology as before (refer to Section~\ref{sec:baseline} and Section~\ref{sec:sim_env}). In total, the train dataset is composed of $1200$ tasks ($2400$ task instances total). Test tasks are also sampled in the same fashion as before, except using the $4$ new objects. Our method is able to succeed at the object picking stage of the tasks $50 \pm 9.9\%$ of the time which is $\sim 2$x better than the best baseline (contextual-LSTM) which only picks $23 \pm 8.4\%$ of the time. Unfortunately, all methods (including ours) often place objects in the wrong bin resulting in final success rates of $23 \pm 8.4\%$ for our method and $22 \pm 8.3\%$ for the best baseline. In practice, this failure mode is easy to rectify since a hard coded policy will always place the object in the right bin. Encouragingly, our policy is best at grasping and picking unseen objects which is the \textit{hardest} part of this task. Nonetheless, this failure mode shows more improvements are needed for this method to work in broader settings.

 \newpage
\subsection{Regularization Effect on Behavior Cloning Loss}
\label{app:loss}
While the inverse model regularization term clearly changed test time performance for the better (shown in Section~\ref{sec:LFA}), can this be explained by positive transfer to the behavior cloning task? In other words, it is possible the inverse modelling loss merely prevents over-fitting in the behavior cloning loss, and thus some other regularization term could achieve the same effect. 

\begin{figure}
    \centering
    
    \includegraphics[width=\columnwidth]{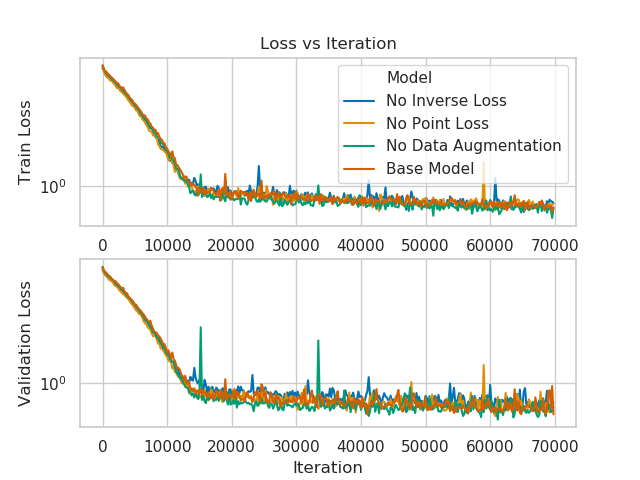}
        
    \captionof{figure}{One hypothesis is that the ablated models fail at test time because they cannot optimize the behavior cloning loss. Comparing train and val loss for models trained on the same data (N=1600) eliminates this possibility.}
    \label{fig:bctrain_curve}
    
\end{figure}

 To test this theory, we plot behavior cloning loss (both training and validation) vs train iteration for both the base model, and ablation models from Section ~\ref{sec:LFA}. Note that behavior cloning train performance is nearly identical, whereas final success rates are dramatically different. We believe these facts in tandem confirm that self-supervised inverse modeling forces our representation to capture information which is useful for robust test performance, but not necessary to minimize the cloning loss. 

 \newpage
 \subsection{Time-Step Ablation}
 \label{app:tstep}
 Instead of using a context video from the demonstrator agent to infer the task, our model could just use the last frame from the demonstration video. After all, the last frame should uniquely specify which object should go in which bin, and prior work~\citep{fu2018variational} has successfully used goal image conditioning. To test this, we train a version of our model which conditions just on the final frame from the context video, and compare its performance on the benchmarks from Section~\ref{sec:baseline}. This modified model achieves a final success rate of $61 \pm 9.7\%$ which is significantly less than the $88 \pm 5.0\%$ our model (which ingests more frames from context) can achieve. This effect holds even if the base model uses just one extra context frames (i.e. both beginning and end frame). We hypothesize that these frames, while not strictly necessary, help the infer which task it needs to perform, thus resulting in a performance boost.

\end{document}